\documentclass[10pt,twocolumn,letterpaper]{article}

\usepackage[pagenumbers]{cvpr} 

\usepackage{graphicx}
\usepackage{amsmath}
\usepackage{amssymb}
\usepackage{booktabs}

\usepackage[numbers,sort]{natbib}
\usepackage{graphicx}
\usepackage{mathtools}
\usepackage{tabularx}
\usepackage{multirow}
\usepackage{enumitem}
\usepackage{bbm}
\usepackage{xcolor, colortbl}
\usepackage{wrapfig}
\usepackage{adjustbox}
\usepackage[accsupp]{axessibility}

\usepackage{physics}

\usepackage{subfiles}
\usepackage{afterpage}

\usepackage[pagebackref,breaklinks,colorlinks]{hyperref}
\definecolor{amethyst}{rgb}{0.6, 0.4, 0.8}

\definecolor{grey}{rgb}{0.9, 0.9, 0.9}
\newcommand{\ccol}{\cellcolor{grey}}
\usepackage{pifont}         

\def\ie{\emph{i.e.}}
\def\eg{\emph{e.g.}}
\def\etal{\emph{et al.}}

\definecolor{grn}{rgb}{0.1, 0.6, 0.1}
\definecolor{mgt}{rgb}{0.8, 0.1, 0.8}
\definecolor{darkblue}{rgb}{0.2, 0.2, 0.8}
\newcommand{\suha}[1]{{\color{darkblue}{#1}}}

\newcommand{\sy}[1]{{\color{purple}{#1}}}

\newcommand{\expnum}[2]{{#1}\mathrm{e}{-#2}}

\makeatletter
\newcommand{\multiline}[1]{%
  \begin{tabularx}{\dimexpr\linewidth-\ALG@thistlm}[t]{@{}X@{}}
    #1
  \end{tabularx}
}
\makeatother

\usepackage[capitalize]{cleveref}
\crefname{section}{Sec.}{Secs.}
\Crefname{section}{Section}{Sections}
\Crefname{table}{Table}{Tables}
\crefname{table}{Tab.}{Tabs.}

\def\cvprPaperID{01957} 
\def\confName{CVPR}
\def\confYear{2023}

\begin{document}

\title{HIER: Metric Learning Beyond Class Labels via Hierarchical Regularization}

\author{
Sungyeon Kim$^1$ \qquad
Boseung Jeong$^1$ \qquad
Suha Kwak$^{1,2}$ \\
Dept. of CSE, POSTECH$^1$ \ \ \ \ \ \ \ \ \ \ \ \ \ \ Graduate School of AI, POSTECH$^2$ \\
{\tt\small \{sungyeon.kim, boseung01, suha.kwak\}@postech.ac.kr} \\
\vspace{-1mm}
{\tt\small \url{http://cvlab.postech.ac.kr/research/HIER}}
\vspace{-1mm}
}
\maketitle


\begin{abstract}
Supervision for metric learning has long been given in the form of equivalence between human-labeled classes.
Although this type of supervision has been a basis of metric learning for decades, we argue that it hinders further advances in the field. 
In this regard, we propose a new regularization method, dubbed HIER, to discover the latent semantic hierarchy of training data, and to deploy the hierarchy to provide richer and more fine-grained supervision than inter-class separability induced by common metric learning losses.
HIER achieves this goal with no annotation for the semantic hierarchy but by learning hierarchical proxies in hyperbolic spaces.
The hierarchical proxies are learnable parameters, and each of them is trained to serve as an ancestor of a group of data or other proxies to approximate the semantic hierarchy among them.
HIER deals with the proxies along with data in hyperbolic space since the geometric properties of the space are well-suited to represent their hierarchical structure.
The efficacy of HIER is evaluated on four standard benchmarks, where it consistently improved the performance of conventional methods when integrated with them, and consequently achieved the best records, surpassing even the existing hyperbolic metric learning technique, in almost all settings.

\end{abstract}

\vspace{-3mm}

\section{Introduction}
\label{sec:intro}



Learning a discriminative and generalizable embedding space has been a vital step within
many machine learning tasks including content-based image retrieval~\cite{kim2019deep,movshovitz2017no,Sohn_nips2016,songCVPR16}, face  verification~\cite{liu2017sphereface,Schroff2015}, person re-identification~\cite{Chen_2017_CVPR,xiao2017joint}, few-shot learning~\cite{Qiao_2019_ICCV, snell2017prototypical, sung2018learning}, and representation learning~\cite{kim2019deep,Wang2015,Zagoruyko_CVPR_2015}. 
Deep metric learning has aroused lots of attention as an effective tool for this purpose. 
Its goal is to learn an embedding space where semantically similar samples are close together and dissimilar ones are far apart.
Hence, the semantic affinity between samples serves as the main supervision for deep metric learning, and has long been given in the form of equivalence between their human-labeled classes.


Although this type of supervision has been a basis of metric learning for decades, we argue that it now hinders further advances of the field. 
The equivalence of human-labeled classes deals with only a tiny subset of possible relations between samples due to the following two reasons.
First, the class equivalence is examined at only a single fixed level of semantic hierarchy, although different classes can be semantically similar if they share the same super-class.
Second, the equivalence is a binary relation that ignores the degree of semantic affinity between two classes.
It is difficult to overcome these two limitations since the semantic hierarchy of data is latent and only human-labeled classes are available from existing datasets in the standard metric learning setting. 
However, once they are resolved, one can open up the possibility of providing rich supervision beyond human-labeled classes. 

\begin{figure} [!t]
\centering
\vspace{-2mm}
\includegraphics[width = 0.95\columnwidth]{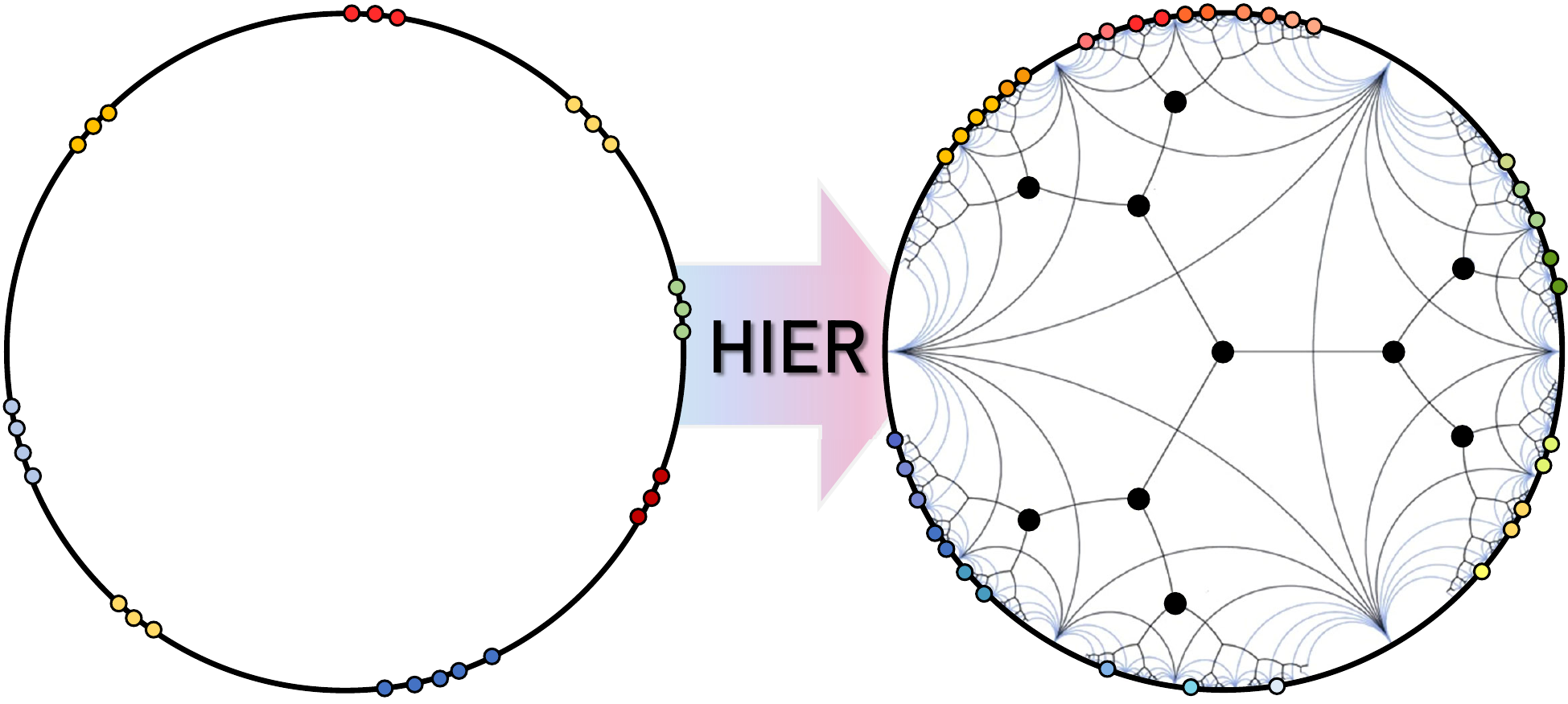}
\caption{
Motivation of HIER. HIER aims to discover an inherent but latent semantic hierarchy of data (colored dots on the boundary) by learning hierarchical proxies (larger black dots) in hyperbolic space. 
The semantic hierarchy is deployed to provide rich and fine-grained supervision that cannot be derived only by human-labeled classes.
} 
\label{fig:thumbnail}
\vspace{-4mm}
\end{figure}

In this regard, we propose a new regularization method, called HIErarchical Regularization (HIER), to discover and deploy the latent semantic hierarchy of training data for metric learning. 
Since the semantic hierarchy can capture not only the predefined set of classes but also their sub-classes, super-classes, and affinities between them, our regularizer is expected to provide richer and more fine-grained supervision beyond inter-class separability induced by common metric learning losses.
However, it is challenging to establish such a semantic hierarchy of data given their human-labeled classes only due to the absence of annotation for the semantic hierarchy.

HIER tackles this challenge by learning \emph{hierarchical proxies} in \emph{hyperbolic space}.
The hierarchical proxies are learnable parameters, and each of them is trained to serve as an ancestor of a group of data or other proxies to approximate the semantic hierarchy. 
Also, HIER deals with the proxies in hyperbolic space since the space is well-suited to represent hierarchical structures of the proxies and data.
It has been reported in literature that Euclidean space with zero curvature is not optimal for representing data exhibiting such a semantic hierarchy~\cite{khrulkov2020hyperbolic}.
In contrast, hyperbolic space with constant negative curvature can represent the semantic hierarchy of data effectively using relatively small dimensions since its volume increases exponentially as its Poincar\'e radius increases~\cite{sarkar2011low, sala2018representation}.

To be specific, HIER is designed as a soft approximation of hierarchical clustering~\cite{wang2018improved, monath2019gradient, dasgupta2016cost}, where similar data should be near by one another in a tree-structured hierarchy while dissimilar ones should be placed in separate branches of the hierarchy (See Figure~\ref{fig:thumbnail}).
During training, HIER constructs a triplet of samples or proxies, in which two of them are similar and the other is dissimilar based on their hyperbolic distances.
Then, the proxy closest to the entire triplet is considered as the \emph{lowest common ancestor} (LCA) of the triplet in a semantic hierarchy; likewise, we identify another proxy as the LCA of only the similar pair in the same manner.
Given the triplet and two LCAs (proxies), HIER 
encourages that each of the LCAs and its associated members of the triplet are close together and that the dissimilar one of the triplet is far apart from the LCA of the similar pair. 
This allows the hierarchical proxies to approximate a semantic hierarchy of data in the embedding space, without any off-the-shelf module for pseudo hierarchical labeling~\cite{yan2021unsupervised}.

The major contribution of our work from the perspective of conventional metric learning is three-fold:
\begin{itemize}
\item We study a new and effective way of providing semantic supervision beyond human-labeled classes, which has not been actively studied in metric learning~\cite{milbich2020diva,roth2019mic, Ge2018DeepML, yang2022hierarchical, yan2021unsupervised}.
\item HIER consistently improved performance over the state-of-the-art metric learning losses when integrated with them, and consequently achieved the best records in almost all settings on four public benchmarks. 
\item By taking advantage of hyperbolic space, HIER substantially improved performance particularly on low-dimensional embedding spaces.
\end{itemize}
%
Also, when regarding our work as a hyperbolic metric learning method, a remarkable contribution of HIER is that it allows taking full advantage of both hyperbolic embedding space and the great legacy of conventional metric learning since it can be seamlessly incorporated with any metric learning losses based on spherical embedding spaces.
The only prior studies on hyperbolic metric learning~\cite{ermolov2022hyperbolic, yan2021unsupervised} are not compatible with conventional losses for metric learning since they learn a distance metric directly on hyperbolic space, in which, unlike the spherical embedding spaces, norms of embedding vectors vary significantly.

\vspace{-1.5mm}
\section{Related Work}
\label{sec:relatedwork}

\subsection{Deep Metric Learning} 
Deep metric learning aims to learn an embedding space where data points of the same class are close to each other, and those of different classes are far apart. To this end, a number of studies have proposed various loss functions, which can be categorized into two ways, pair-based and proxy-based losses.
Pair-based losses handle the relations between pairs of data points. Contrastive loss aims to minimize the distance between positive pairs and maximize the distance between negative pairs. Triplet loss~\cite{Schroff2015, Wang2014} employs a triplet of anchor, positive and negative samples, and optimizes the embedding space by enforcing the constraint that the distance between the positive pair is smaller than that between the negative pair with a pre-defined margin. Higher-order variants of these losses have also been proposed to capture more complex relations between embedding vectors~\cite{Sohn_nips2016, songCVPR16, wang2019ranked, wang2019multi}.
On the other hand, proxy-based losses consider the relations between data points and proxies, where the proxy is an additional learnable embedding vector that represents the class of training data. Proxy-NCA~\cite{movshovitz2017no} associates a data point with proxies and pulls the positive pair of a data point and proxy closer while pushing away the negative pair. Proxy-Anchor~\cite{kim2020proxy} associates a proxy with all data points in a batch, allowing it to consider rich relations between data points while reflecting their relative hardness through gradients.

Recently, loss functions that leverage the underlying hierarchical relations of data beyond class equivalence relations have been proposed for metric learning, such as hierarchical triplet loss~\cite{Ge2018DeepML} and hierarchical proxy-based loss~\cite{yang2022hierarchical}. These methods predefine the hierarchy of data, and adjust the distances between data points to fit the discrete hierarchy. In contrast, HIER aims to learn \emph{continuous hierarchical representations} in a data-driven manner without the need for prior knowledge of the hierarchy or pre-existing algorithms~\cite{yan2021unsupervised}. This is a significant advantage over existing hierarchical metric learning methods~\cite{Ge2018DeepML, yang2022hierarchical, yan2021unsupervised}, which have a discrete nature and require the number of hierarchy levels and clusters per level to be specified. Our method allows for a more robust and adaptable method for hierarchical clustering, automatically adjusting to the latent semantic hierarchy of the data.

\subsection{Hyperbolic Embedding} 
Learning hyperbolic embedding has drawn the attention of many research fields because it can encode data such as text or images into hyperbolic space with semantically rich representation due to its high capacity and tree-like property.
Since learning embedding in the hyperbolic space for natural language processing~\cite{nickel2017poincare} has been successful, there are some attempts to learn hyperbolic embeddings for computer vision tasks such as few-shot learning~\cite{khrulkov2020hyperbolic, gao2021curvature} and deep metric learning~\cite{ermolov2022hyperbolic, yan2021unsupervised}; they have demonstrated that hyperbolic embeddings are able to improve the model performance.
Instead of building complex network architectures whose all layers operate in hyperbolic space, they proposed hybrid architectures where only the last layer maps inputs in Euclidean space to the hyperbolic embedding vectors, and the remainders still operate in Euclidean space.
Yan~\etal~\cite{yan2021unsupervised} proposed the unsupervised hyperbolic metric learning framework by conducting hierarchical clustering with pre-defined hierarchy thresholds.
Ermolov~\etal~\cite{ermolov2022hyperbolic} utilized the pairwise cross-entropy loss with hyperbolic distances through joint use of vision transformers~\cite{ViT, DeiT, DINO}.

Unfortunately, these approaches do not fully take advantage of hyperbolic space for the following reasons.
First, a pre-defined set of hierarchy thresholds restricts the hierarchical property of hyperbolic space.
Second, the learned embedding space is less generalizable since the latent semantic hierarchy of data in hyperbolic space, which can provide more fine-grained supervision and allow a more generalizable embedding space to be learned, is not considered.
Contrary to the existing hyperbolic metric learning methods, our loss can preserve the heritage of metric learning as well as fully take advantage of hyperbolic space thanks to HIER.


\subsection{Hierarchical Clustering}
Hierarchical clustering is a recursive grouping of a dataset into clusters with an increasingly finer granularity.
It has usually been used for data analysis~\cite{zhao2002evaluation, sorlie2001gene}, visualization~\cite{seo2002interactively}, and mining fine-grained relations of data~\cite{brown1992class}.
Recently, cost function based methods~\cite{dasgupta2016cost, wang2018improved, monath2019gradient} have been developed.

Dasgupta~\cite{dasgupta2016cost} first proposed a cost function based method, where the cost function is optimized by pairwise similarities between data points and the leaves of the lowest common ancestor (LCA).
Wang and Wang~\cite{wang2018improved} have improved the cost function of~\cite{dasgupta2016cost} by introducing a triplet manner that determines the weights of the costs according to the triplet relations.
However, These cost functions cannot be optimized by the stochastic gradient descent method due to its inherent discrete nature.
As an effort to overcome this limitation, Monath~\etal~\cite{monath2019gradient} proposed a gradient-based hyperbolic hierarchical clustering (gHHC) that can be applied to continuous tree representations in hyperbolic space, and improved computational cost by approximating a distribution over LCA for two or three data points.

The aforementioned works have enabled hierarchical clustering for both discrete and continuous trees, but they have the limitation of requiring a ground-truth weighted graph. Therefore, they are not suitable for hyperbolic metric learning benchmarks, where the ground-truth weighted graph is absent.
HIER can overcome this problem by leveraging virtual ancestors of data points, named as \textit{hierarchical proxies}, while still allowing for stochastic gradient descent optimization.

\vspace{-2mm}

\section{Proposed Method}
\label{sec:method}
This section first presents preliminaries to our work, the Poincar\'e ball model and hierarchical clustering. 
Then, the objective of HIER is elaborated on. 

\subsection{Preliminary}
\subsubsection{The Poincar\'e ball Model}
Hyperbolic space means a Riemannian manifold with negative curvature, which has multiple conformal models~\cite{cannon1997hyperbolic}. In this work, we use the Poincar\'e ball, which is a popular and well-studied space in hyperbolic geometry. 
The $n$-dimensional Poincar\'e ball model $(\mathbb{D}_c^n, g^\mathbb{D})$ is defined by the manifold $\mathbb{D}^n_c = \{\vb{x} \in \mathbb{R}^n: c||\vb{x}|| < 1 \}$ and Riemannian metric $g^\mathbb{D} = {\lambda^2_{c}}{g^{E}}$, where $c$ is a curvature hyperparameter, $\lambda_c = \frac{2}{1-c||\vb{x}||^2}$ is the conformal factor, and $g^{E} = I_n$ is Euclidean metric tensor. 

An output of a conventional embedding network, lying in Euclidean space, can be transformed into a point on a Poincar\'e ball by a mapping function called \emph{exponential map}.
The common form of the exponential map is given by
\begin{equation}
\exp_0(\vb{v}) = \tanh{\sqrt{c}||\vb{v}||}\frac{\vb{v}}{\sqrt{c}||\vb{v}||}.
\label{eq:exp_map}
\end{equation}
One of the important properties of hyperbolic space is that it is not a vector space, so vectors can be calculated algebraically only by introducing gyrovector spaces~\cite{ungar2008gyrovector}, a generalized notion of vector spaces.
The addition operation in gyrovector spaces, called  M\"obius addition, between a pair of vectors $\vb{u} \in \mathbb{D}_c^n$ and $\vb{v} \in \mathbb{D}_c^n$ is defined as
\begin{equation}
\vb{u}{\oplus_c}\vb{v}:=\frac{(1+2c\langle{\vb{u}},\vb{v}\rangle+c||\vb{v}||^2)\vb{u}+(1-c||\vb{u}||^2)\vb{v}}
{1+2c\langle{\vb{u}},\vb{v}\rangle+c^2||\vb{u}||^2||\vb{v}||^2}.
\label{eq:mobius_addition}
\end{equation}
In the Poincar\'e ball, the hyperbolic distance between the two vectors $\vb{u}$ and $\vb{v}$ is then formulated as
\begin{equation}
d_{H}(\vb{u},\vb{v}) = \frac{2}{\sqrt{c}}\mathrm{arctanh}\big(\sqrt{c}||-\vb{u}{\oplus_c}\vb{v}||\big).
\label{eq:hyperbolic_distance}
\end{equation}
As the curvature $c$ of Eq.~\eqref{eq:hyperbolic_distance} approaches to $0$,
the hyperbolic distance becomes equal to the Euclidean distance. 
As the norm of vectors increases, the hyperbolic distance grows much faster than the Euclidean distance that increases linearly.
Thanks to its inherent characteristic, when the radius of the Poincar\'e ball increases, its volume increases exponentially and thus enables to represent the semantic hierarchy of data even with low dimensions~\cite{sarkar2011low, sala2018representation}. 
However, at the same time, this property leads to serious optimization issues when the hyperbolic distance is incorporated with existing metric learning losses.
Specifically, the range of hyperbolic distances is not bounded, and accordingly their scales vary significantly by norms of embedding vectors. Thus, the use of hyperbolic distance makes the criteria of conventional metric learning losses futile.
One tentative solution is to normalize the embedding vectors and utilize an existing metric learning loss. 
However, such a na\"ive method cannot encode hierarchical structures, which is the main reason for adopting hyperbolic space.

\vspace{-3mm}
\subsubsection{Conventional Hierarchical Clustering}
Hierarchical clustering is an algorithm that builds a hierarchy of clusters in a tree-like structure by grouping data with a progressively finer granularity.
Intuitively, the optimal clustering constructs a tree where similar data points are close to each other and dissimilar data points are located in separate branches.
The Dasgupta cost~\cite{dasgupta2016cost} is an objective function that measures the quality of hierarchical clustering based on this property, and is formulated as follows:

\begin{equation}
\mathcal{C} := \sum_{i,j}w_{i,j}|\mathrm{leaves}(i\vee{j})|,
\label{eq:dasgupta_cost}
\end{equation}
where $w_{i,j}$ is the ground-truth weight indicating similarity between two nodes $i$ and $j$, $i\vee{j}$ denotes their lowest common ancestor~(LCA), and $\mathrm{leaves}(k)$ is a set of leaves of the sub-tree whose root is node $k$. 
To minimize the Dasgupta cost, the higher the ground-truth weight between two nodes, the smaller the number of descendants of their LCA should be. This means that similar nodes should be located close to each other at leaves of the tree structure.

Unfortunately, this cost function cannot be optimized by stochastic gradient descent because of its discrete nature.
Its extensions~\cite{chami2020trees, monath2019gradient} enable gradient-based optimization for clustering, but they still have limitations in that they demand the ground-truth weighted graph, which is not available in metric learning datasets. 
For these reasons, the Dasgupta cost and its extensions are not directly applicable to deep metric learning.

\begin{figure*} [!t]
\centering
\vspace{-3mm}
\includegraphics[width = 0.9 \linewidth]{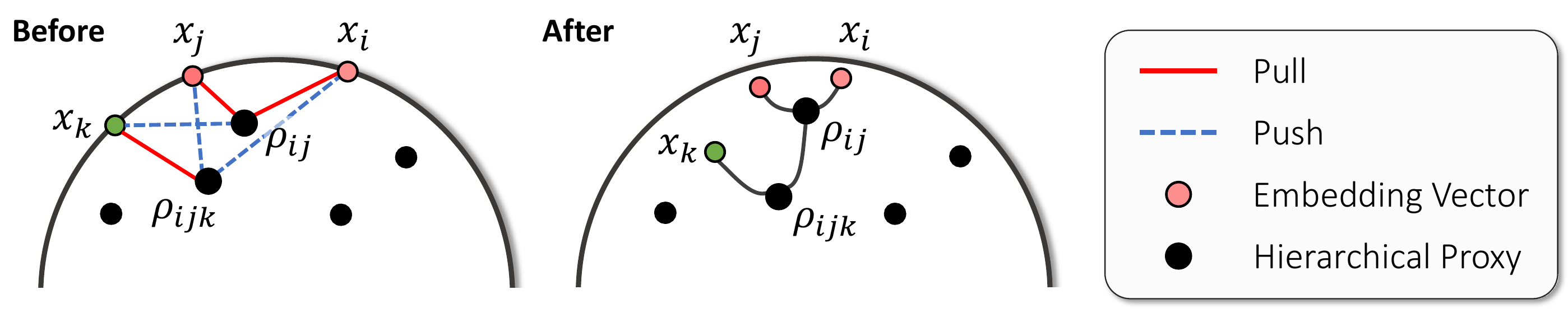}
\vspace{-3mm}
\caption{
A conceptual illustration of the learning objective in Eq.~\eqref{eq:hc_loss}. Each hierarchical proxy is colored in black and different colors indicate different classes. The associations defined by the losses are expressed by edges, where the red solid line means the pulling and the blue dotted line is the pushing. Relevant pairs are pulled into their LCA, and the remaining sample is pulled into LCA of the entire triplet.
} 
\label{fig:our_reg}
\vspace{-3mm}
\end{figure*}

\subsection{Our Hierarchical Metric Learning Objective}

We proposes HIER, a novel regularization method that discovers and deploys the latent semantic hierarchy of data in hyperbolic space.
Since the semantic hierarchy is expected to capture not only predefined classes but also their sub-classes, super-classes, and affinities between these notions, our regularizer provides more fine-grained supervision beyond inter-class separability induced by common metric learning losses. 
Also, it is designed to be incorporated with any conventional metric learning losses based on the spherical embedding.



\vspace{-3mm}
\subsubsection{HIER}

To produce rich supplementary supervision as a regularizer, HIER aims to discover the latent semantic hierarchy of data with no additional annotation for the hierarchy.
Specifically, it is built upon an extension of the Dasgupta cost~\cite{monath2019gradient}, but is reshaped for unsupervised clustering with no ground-truth weighted graph.
The key idea of HIER is to employ \emph{hierarchical proxies} as learnable ancestors of data points in the hierarchy.

Given a triplet $\{ x_i, x_j, x_k \}$, in which $x_i$ and $x_j$ are related to each other and $x_k$ is irrelevant, HIER encourages that the pair of related samples have the same LCA and the rest has a different LCA.
Note that we do not consider class labels of data when sampling such a triplet since the goal of HIER is to discover the latent semantic hierarchy beyond the predefined classes.
Instead, we employ the reciprocal nearest neighbor to ensure the relevance relations of a triplet.
To be specific, the set of feasible triplets, denoted by $\mathcal{T}$, is estimated as follows:
\begin{align}
\begin{split}
    &\mathcal{T} = \{(x_i,x_j,x_k) \ | \  (x_j \in R_K(x_i)) \wedge (x_k \not\in R_K(x_i))\} \\ 
    &\hspace{-0.5em}\textrm{where } R_K(x) = \{x'|(x' \in N_K(x))\wedge (x \in N_K(x'))\},
\end{split}
\label{eq:triplet_condition}
\end{align}
where $x \in \mathbb{D}_c^n$ denotes an embedding vector in hyperbolic space, $N_K(x)$ is the $K$-nearest neighbors of $x$, and $R_K(x)$ stands for the $K$-reciprocal nearest neighbors of $x$.
The probability that a hierarchical proxy $\rho \in P$ is the LCA of embedding vectors $x_i$ and $x_j$ is given by
\begin{equation}
    \pi_{ij} (\rho) = \exp\Big(- \max\big\{d_H(x_i,\rho), d_H(x_j,\rho)\big\}\Big),
\label{eq:lca_prob}
\end{equation}
where $d_H$ is the hyperbolic distance defined in Eq.~\eqref{eq:hyperbolic_distance}.
Then, the most likely LCA of $x_i$ and $x_j$ is sampled from the probability distribution $\pi_{ij}$ by the Gumbel-max trick as follows:
\begin{align}
\rho_{ij} = \underset{\rho}{\arg\max}\big(\pi_{ij}(\rho) + g_{ij}\big),
\label{eq:lca_sampling}
\end{align}
where $g_{ij}$ is an \emph{i.i.d.} sample drawn from $\textrm{Gumbel}(0,1)$;
the Gumbel noise $g_{ij}$ allows to avoid falling into local optima.
In the same way of Eq.~\eqref{eq:lca_prob} and \eqref{eq:lca_sampling}, $\rho_{ijk}$ that is likely to be the LCA of the entire triplet $\{ x_i, x_j, x_k \}$ is sampled from the hierarchical proxy set $P$ except for $\rho_{ij}$.

Using the hyperbolic distances between a sampled triplet and their LCAs, our objective for hierarchical regularization is implemented as a stack of three triplet losses:
\begin{equation}
\begin{split}
\mathcal{L}_{\textrm{HIER}}(t) &= [d_H(x_i, \rho_{ij}) - d_H(x_i, \rho_{ijk}) + \delta]_+ \\
&+ [d_H(x_j, \rho_{ij}) - d_H(x_j, \rho_{ijk}) + \delta]_+ \\
&+ [d_H(x_k, \rho_{ijk}) - d_H(x_k, \rho_{ij}) + \delta]_+,
\end{split}
\label{eq:hc_loss}
\end{equation}
where $\delta$ is a margin hyperparameter.
Figure~\ref{fig:our_reg} illustrates the behavior of HIER in terms of the relations between a triplet and their LCAs. HIER forces the relevant samples $x_i$ and $x_j$ to be close to $\rho_{ij}$, but encourages them to increase the distance from $\rho_{ijk}$ by the margin. Meanwhile, the opposite signal is applied to the irrelevant sample $x_k$. These operations encourage $\rho_{ijk}$ to be relatively close to the center of the Poincar\`e ball and $\rho_{ij}$ to be away from the center. Consequently, they become representatives of higher and lower levels of the semantic hierarchy, respectively. In addition, relevant samples are clustered, while unrelated samples are branched out from them.
As a result, $x_i$ and $x_j$ belong to a child of $\rho_{ij}$, and $x_k$ becomes a child of $\rho_{ijk}$, forming a tree-like hierarchical structure.


\subsubsection{Total Objective}
Although HIER provides rich and fine-grained supervision induced by the approximate semantic hierarchy of training data, a metric learning loss still plays crucial roles as its supervisory signals are reliable thanks to the use of ground-truth labels while HIER relies on self-supervised hierarchical clustering.
The final loss function for metric learning is thus a linear combination of the two terms as follows:
\begin{equation}
\mathcal{L} = \mathcal{L}_{\textrm{ML}} + \lambda \sum_{t \in \{\mathcal{T}_x,\mathcal{T}_\rho\}}\mathcal{L}_{\textrm{HIER}}(t),
\label{eq:total_loss}
\end{equation}
where $\lambda$ is a weight hyperparameter, $\mathcal{L}_{\textrm{ML}}$ is a metric learning loss, and $\mathcal{T}_x$ and $\mathcal{T}_\rho$ are sets of triplets of samples and hierarchical proxies that satisfy the condition in Eq.~\eqref{eq:triplet_condition}, respectively. By feeding both samples and proxies as input to the loss, we imply $\mathcal{L}_{\textrm{HIER}}$ to encourage hierarchical relations between ancestors (\ie, proxies) as well as individual samples.

For the metric learning objective $\mathcal{L}_{\textrm{ML}}$, either those based on cosine~(Euclidean) distances or those of hyperbolic distances can be incorporated. When using a metric learning loss based on spherical embedding, the loss is calculated through the Euclidean metric with $\ell_2$ normalized embedding vectors. 
In this case, the metric learning loss only controls the angles between the normalized embedding vectors and HIER adjusts their hyperbolic distances based on their positions and norms.






\section{Experiments}
\label{sec:experiments}
In this section, we evaluate our method on four benchmark datasets for deep metric learning and compare it with the state of the art.

\begin{table*}[!t]
\fontsize{8.1}{10.0}\selectfont
\centering
\begin{tabularx}{0.99\textwidth}
    {
      p{0.12\textwidth}
      >{\centering\arraybackslash}X
      >{\centering\arraybackslash}X
      >{\centering\arraybackslash}X 
      >{\centering\arraybackslash}X
      >{\centering\arraybackslash}X
      >{\centering\arraybackslash}X
      >{\centering\arraybackslash}X 
      >{\centering\arraybackslash}X
      >{\centering\arraybackslash}X
      >{\centering\arraybackslash}X
      >{\centering\arraybackslash}X
      >{\centering\arraybackslash}X
      >{\centering\arraybackslash}X
      }
     \toprule
    \multicolumn{1}{l}{\multirow{2}{*}[-3.5mm]{Methods}}&
    \multicolumn{1}{c}{\multirow{2}{*}[-3.5mm]{Arch.}}&
    \multicolumn{3}{c}{CUB} & \multicolumn{3}{c}{Cars} & \multicolumn{3}{c}{SOP}& \multicolumn{3}{c}{In-Shop}\\ 
    \cmidrule(lr){3-5} \cmidrule(lr){6-8} \cmidrule(lr){9-11} \cmidrule(lr){12-14}
    & &R@1 &R@2 &R@4 &R@1 &R@2 &R@4 &R@1 &R@10 &R@100  &R@1 &R@10 &R@20 \\ \midrule
    \multicolumn{14}{l}{{\footnotesize \textit{Backbone architecture}: \textit{CNN}}} \\ \midrule
    NSoftmax~\cite{zhai2018classification} &\multicolumn{1}{l}{R$^{128}$} &56.5 &69.6 &79.9  &81.6 &88.7 &93.4  &75.2 &88.7 &95.2&86.6 &96.8 &97.8 \\
    MIC~\cite{roth2019mic} &\multicolumn{1}{l}{R$^{128}$} &66.1 &76.8 &85.6&82.6 &89.1 &93.2 &77.2 &89.4 &94.6 &88.2 &97.0 &- \\
    XBM~\cite{wang2020cross} &\multicolumn{1}{l}{R$^{128}$} &- &- &- &- &- &- &80.6 &91.6 &96.2  &91.3 &97.8 &98.4\\ 
    \midrule
    XBM~\cite{wang2020cross} &\multicolumn{1}{l}{B$^{512}$} &65.8 &75.9 &84.0 &82.0 &88.7 &93.1 &79.5 &90.8 &96.1 &89.9 &97.6 &98.4  \\
    HTL~\cite{Ge2018DeepML} &\multicolumn{1}{l}{B$^{512}$} &57.1 &68.8 &78.7 &81.4 &88.0 &92.7 &74.8 &88.3 &94.8 &80.9 &94.3 &95.8  \\
    MS~\cite{wang2019multi} &\multicolumn{1}{l}{B$^{512}$} &65.7 &77.0 &86.3 &84.1 &90.4 &94.0 &78.2 &90.5 &96.0 &89.7 &97.9 &98.5  \\
    SoftTriple~\cite{Qian_2019_ICCV} &\multicolumn{1}{l}{B$^{512}$} &65.4 &76.4 &84.5 &84.5 &90.7 &94.5 &78.6 &86.6 &91.8 &- &- &- \\
    PA~\cite{kim2020proxy} &\multicolumn{1}{l}{B$^{512}$} &68.4 &79.2& 86.8 &86.1 & 91.7 & 95.0 &79.1 & 90.8 & 96.2 &91.5 & 98.1 & 98.8 \\
    NSoftmax~\cite{zhai2018classification} &\multicolumn{1}{l}{R$^{512}$} &61.3 &73.9 &83.5 &84.2 &90.4 &94.4 &78.2 &90.6 &96.2 &86.6 &97.5 &98.4 \\
    $^\dagger$ProxyNCA++~\cite{teh2020proxynca++} &\multicolumn{1}{l}{R$^{512}$} &69.0 &79.8 &87.3 &86.5 &92.5 &95.7 &80.7 &92.0 &96.7 &90.4 &98.1 &98.8 \\
    Hyp~\cite{ermolov2022hyperbolic} &\multicolumn{1}{l}{R$^{512}$} &65.5 &76.2 &84.9 &81.9 &88.8 &93.1 &79.9 &91.5 &96.5 &90.1 &98.0 &98.7  \\
    \ccol HIER (ours) & \multicolumn{1}{l}{\ccol R$^{512}$} & \ccol 70.1 & \ccol 79.4 &\ccol  86.9& \ccol 88.2&\ccol 93.0&\ccol 95.6& \ccol 80.2&\ccol  91.5& \ccol96.6& \ccol 92.4& \ccol98.2& \ccol 98.8 \\ \midrule
    
    
    \multicolumn{14}{l}{{\footnotesize \textit{Backbone architecture}: \textit{ViT}}} \\ \midrule
    IRT$_\text{R}$~\cite{el2021training} &\multicolumn{1}{l}{De$^{128}$} &72.6 &81.9 &88.7 &- &- &- &83.4 &93.0 &97.0  &91.1 &98.1 &98.6 \\
    Hyp~\cite{ermolov2022hyperbolic} &\multicolumn{1}{l}{De$^{128}$} &74.7 &84.5 &90.1 &82.1 &89.1 &93.4 &83.0 &93.4 &97.5  &90.9 &97.9 &98.6  \\
    \ccol HIER (ours) & \multicolumn{1}{l}{\ccol De$^{128}$} & \ccol 75.2 & \ccol 84.2 &\ccol  90.0& \ccol 85.1&\ccol 91.1&\ccol 95.1& \ccol 82.5&\ccol  92.7& \ccol97.0& \ccol 91.0& \ccol98.0& \ccol 98.6 \\
    Hyp~\cite{ermolov2022hyperbolic} &{DN}$^{128}$ &78.3 &86.0 &91.2 &86.0 &91.9 &95.2 &84.6 &94.1 &97.7 &92.6 &98.4 &99.0\\
    \ccol HIER (ours) &\multicolumn{1}{l}{\ccol DN$^{128}$}   & \ccol 78.5& \ccol86.7& \ccol \ccol 91.5&  \ccol 88.4& \ccol 93.3& \ccol 95.9&  \ccol 84.9& \ccol 94.2& \ccol 97.5& \ccol \ccol 92.6& \ccol 98.4& \ccol 98.9 \\ 
    Hyp~\cite{ermolov2022hyperbolic} & \multicolumn{1}{l}{V$^{128}$} &84.0 &90.2 &94.2&82.7 &89.7 &93.9  &85.5 &94.9 &98.1&92.7 &98.4 &98.9 \\
    \ccol HIER (ours) & \multicolumn{1}{l}{\ccol V$^{128}$}  & \ccol 84.2& \ccol 90.1& \ccol 93.7&\ccol 86.4&\ccol  91.9& \ccol 95.1& \ccol 85.6& \ccol 94.6&
    \ccol 97.8& \ccol 92.7& \ccol 98.4& \ccol 98.9 \\ \midrule
    
    IRT$_\text{R}$~\cite{el2021training}  &\multicolumn{1}{l}{De$^{384}$}  &76.6 &85.0 &91.1 &- &- &- &84.2 &93.7 &97.3 &91.9 &98.1 &98.9  \\
    DeiT-S~\cite{DeiT} &\multicolumn{1}{l}{ De$^{384}$}  &70.6 &81.3 &88.7 &52.8 &65.1 &76.2 &58.3 &73.9 &85.9 &37.9 &64.7 &72.1  \\
    Hyp~\cite{ermolov2022hyperbolic} &\multicolumn{1}{l}{ De$^{384}$}  &77.8 &86.6 &91.9 &86.4 &92.2 &95.5 &83.3 &93.5 &97.4 &90.5 &97.8 &98.5 \\
    \ccol HIER (ours) &\multicolumn{1}{l}{\ccol De$^{384}$}   & \ccol 78.7& \ccol 86.8& \ccol 92.0 &  \ccol88.9& \ccol 93.9& \ccol 96.6& \ccol 83.0& \ccol 93.1& \ccol 97.2& \ccol 90.6& \ccol 98.1& \ccol 98.6\\
    DINO~\cite{DINO} &\multicolumn{1}{l}{ DN$^{384}$}  &70.8 &81.1 &88.8 &42.9 &53.9 &64.2 &63.4 &78.1 &88.3 &46.1 &71.1 &77.5 \\
    Hyp~\cite{ermolov2022hyperbolic} &\multicolumn{1}{l}{ DN$^{384}$}  &80.9 &87.6 &92.4  &89.2 &94.1 &96.7 &85.1 &94.4 &97.8 &92.4 &98.4 &98.9 \\
    \ccol HIER (ours) & \multicolumn{1}{l}{\ccol DN$^{384}$} & \ccol 81.1& \ccol 88.2& \ccol 93.3& \ccol 91.3& \ccol 95.2& \ccol 97.1& \ccol 85.7& \ccol  94.6& \ccol 97.8&\ccol  92.5& \ccol 98.6& \ccol 99.0\\
    ViT-S~\cite{ViT} &\multicolumn{1}{l}{V$^{384}$} &83.1 &90.4 &94.4 &47.8 &60.2 &72.2 &62.1 &77.7 &89.0 &43.2 &70.2 &76.7 \\
    Hyp~\cite{ermolov2022hyperbolic} &\multicolumn{1}{l}{V$^{384}$}  &85.6 &91.4 &94.8 &86.5 &92.1 &95.3 &85.9 &94.9 &98.1 &92.5 &98.3 &98.8 \\
    \ccol HIER (ours) & \multicolumn{1}{l}{\ccol  V$^{384}$} & \ccol 85.7& \ccol 91.3& \ccol 94.4& \ccol 88.3& \ccol 93.2& \ccol 96.1& \ccol 86.1& \ccol 95.0& \ccol 98.0& \ccol 92.8& \ccol 98.4& \ccol 99.0\\
    \bottomrule
\end{tabularx}
\vspace{-2mm}
\caption{
Performance of metric learning methods on the four datasets. Their network architectures are denoted by abbreviations, R--ResNet50~\cite{resnet}, B--Inception with BatchNorm~\cite{Batchnorm}, De--DeiT~\cite{DeiT}, DN--DINO~\cite{DINO} and V--ViT~\cite{ViT}. Note that ViT~\cite{ViT} is pretrained on ImageNet-21k~\cite{Imagenet}. Superscripts denote their embedding dimensions and $\dagger$ indicates models using larger input images.}
\label{tab:Hyp_sup_head}
\vspace{-4mm}
\end{table*}

\subsection{Experimental Setup}
\label{subsec:experiment_unsup_detail}
\noindent \textbf{Datasets.}
On four benchmark datasets, namely CUB-200-2011 (CUB)~\cite{CUB200}, Cars-196 (Cars)~\cite{krause20133d}, Stanford Online Product (SOP)~\cite{songCVPR16}, and In-shop Clothes Retrieval (In-Shop)~\cite{DeepFashion}, models are evaluated and compared. 
We split the datasets into train and test sets, directly following the standard protocol presented in \cite{kim2020proxy}.


\vspace{1mm}
\noindent \textbf{Evaluation metric.}
We measure the performance on the datasets by Recall@$k$, the fraction of queries that have at least one relevant sample in their $k$-nearest neighbors on a learned hyperbolic embedding space.


\vspace{1mm}
\noindent \textbf{Embedding network architectures.}
For fair comparisons with previous work, we utilize ResNet50~\cite{resnet} and vision transformer architecture with three types of pretraining scheme (ViT-S~\cite{ViT}, DeiT-S~\cite{DeiT} and DINO~\cite{DINO}). All encoder is pretrained for ImageNet classification~\cite{Imagenet}. In ViT variants, the linear projection layer for patch embedding is frozen during training. We adjust the size of the last FC layer according to the dimensionality of embedding vectors. We note that $L_2$ normalization is not used for hyperbolic embedding. Instead, the exponential mapping layer~(See Eq.~\eqref{eq:exp_map}) is appended to the last embedding layer.

\vspace{1mm}
\noindent \textbf{Implementation details.}
Our embedding models are optimized by AdamW~\cite{adamw} with the learning rate value $10^{-5}$ for ViT-S and DeiT-S, and $5\times10^{-6}$ for DINO models. For experiments using ResNet50, we follow the training setting of \cite{kim2020proxy}.
Training images are randomly resized and cropped to $224 \times 224$ and randomly flipped horizontally while test images are resized to $256 \times 256$ and then center cropped.
For hyperbolic embedding, we use curvature parameter $c= 0.1$  and clipping radius $r=2.3$ following previous work~\cite{ermolov2022hyperbolic}.
In proxy anchor loss, we use a high learning rate for proxies by scaling $1\times10^4$ times. 
For all of our experiments, we maintain a consistent set of hyperparameters, including $K=20$ for the number of nearest neighbors, $|P|=512$ for the number of hierarchical proxies, $\lambda=1$ for the loss weight of HIER, and $\delta = 0.1$ for the margin. 

\vspace{-1mm}
\subsection{Quantitative Results}
We compare the performance of the proposed method with the existing state of the arts on four standard datasets. In these experiments, we adopt proxy anchor loss~\cite{kim2020proxy} as a metric learning objective $\mathcal{L}_\textrm{ML}$. For a fair comparison with other methods, we divide the previous work in terms of backbone architecture and embedding dimension, which is summarized in Table~\ref{tab:Hyp_sup_head}.

\begin{figure*} [!t]
\centering
\vspace{-2mm}
\includegraphics[width = 0.9 \linewidth]{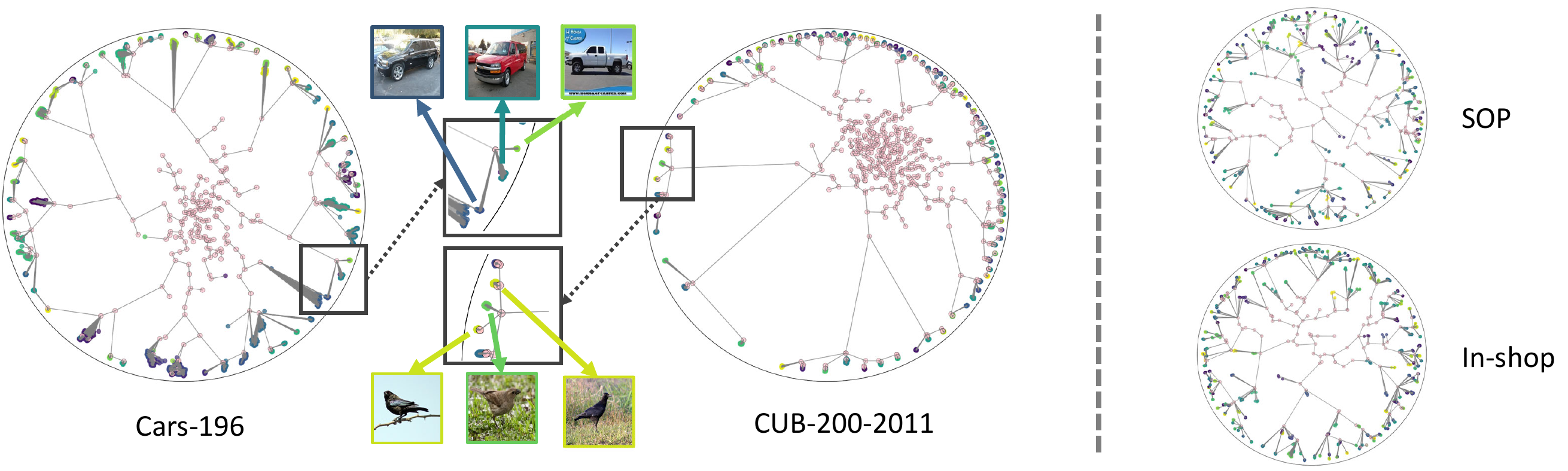}
\vspace{-2mm}
\caption{UMAP visualization of our embedding space learned on the train split of Cars, CUB, SOP, and In-shop. Pink ones indicate hierarchical proxies and other colors represent distinct classes. The gray lines are the ancestor-descendant relations between data points.
} 
\label{fig:visualization_map}
\vspace{-2mm}
\end{figure*}

\begin{figure*} [!t]
\centering
\includegraphics[width = 0.93 \linewidth]{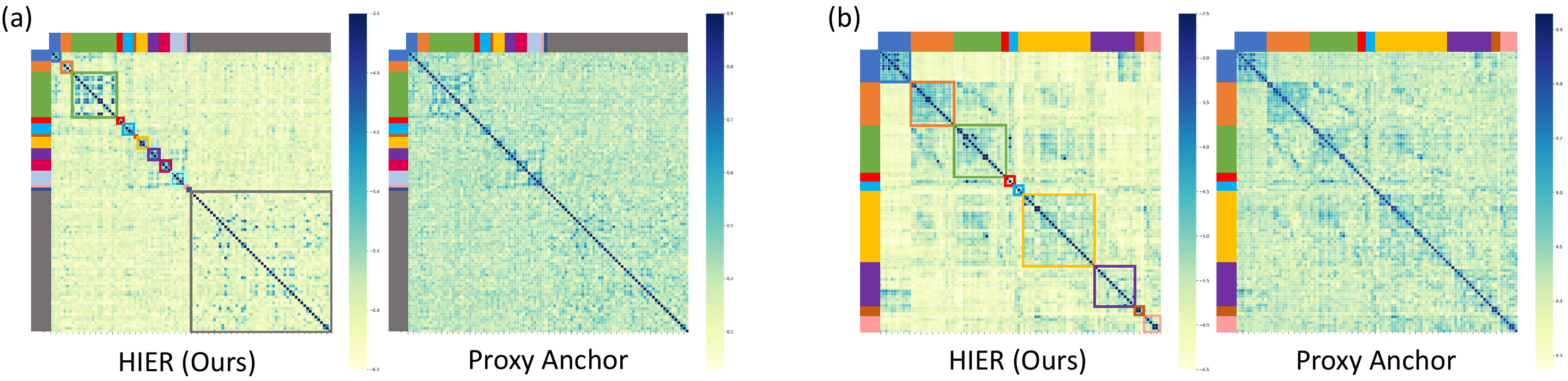}
\vspace{-2mm}
\caption{
Class-to-class affinity matrices of proxy anchor and ours on the CUB (a) and Cars (b) datasets, which show the inter-class similarity. The different colors (13 colors of CUB and 9 colors of Cars) of the sidebar indicate distinct actual super-classes at the order level referring to the hierarchy labels of \cite{chen2022label}.
} 
\label{fig:sup_class}
\vspace{-4mm}
\end{figure*}

The experimental results demonstrate the effectiveness of our proposed method. Compared with existing CNN-based methods, our method using ResNet50 as the backbone achieves superior performance on all datasets except for SOP. Specifically, it outperforms ProxyNCA++ by a significant margin of 1.7\% on the Cars dataset and 2\% on the In-Shop dataset. Furthermore, we reimplement Hyp using ResNet50 and observe that its performance lags behind ours.
For ViT-based models, HIER consistently outperforms the state of the art in terms of R@1 for almost all settings. Specifically, our method with 128 embedding dimensions achieves a large margin improvement of 3.7\%, 2.4\%, and 3.0\% in R@1 for V$^{128}$, DN$^{128}$, and De$^{128}$ models on the Cars dataset, respectively, and surpasses the performance of 512-dimensional CNN-based methods.
These improvements suggest that our method allows the embedding space to be more discriminative and generalized with smaller dimensionality, thanks to utilizing hyperbolic space and considering latent semantic hierarchy.

\subsection{Qualitative Results}
To demonstrate that HIER actually represents a latent semantic hierarchy of data, we visualize the learned embedding vectors which are projected to a 2-dimensional Poincar\'e ball. For visualization, we use UMAP~\cite{mcinnes2018umap} with hyperboloid distance metric as a dimensional reduction technique. Figure~\ref{fig:visualization_map} shows that a tree-like hierarchical structure between data and hierarchical proxies is constructed in the embedding space learned by HIER. The middle figures are enlarged figures of part of the embedding space, which shows that samples in a sub-tree share common semantics, although not of the same class. This visualization proves that HIER discovers and deploys latent and meaningful hierarchies between data in the embedding space.


\subsection{Analysis on Semantic Hierarchy of HIER}
To demonstrate that HIER can capture a meaningful semantic hierarchy, we perform two qualitative experiments that can show whether the embedding space reflects the super-class and sub-class relations between data.

In Figure~\ref{fig:sup_class}, we present the affinity matrix between classes on the embedding space trained with our method, and compare it with that of the proxy anchor loss~\cite{kim2020proxy}. The similarity measures of affinity matrices for proxy anchor loss and ours are set to cosine similarity and negative hyperbolic distance, respectively. Since classes with the same super-class share common attributes, they have a high degree of similarity in both methods.
However, in the case of proxy anchor loss, classes belonging to different super-classes often have high similarity, whereas HIER can clearly distinguish between classes sharing the same super-class and classes that are not. 

\begin{figure} [!t]
\centering
\includegraphics[width = 0.96 \columnwidth]{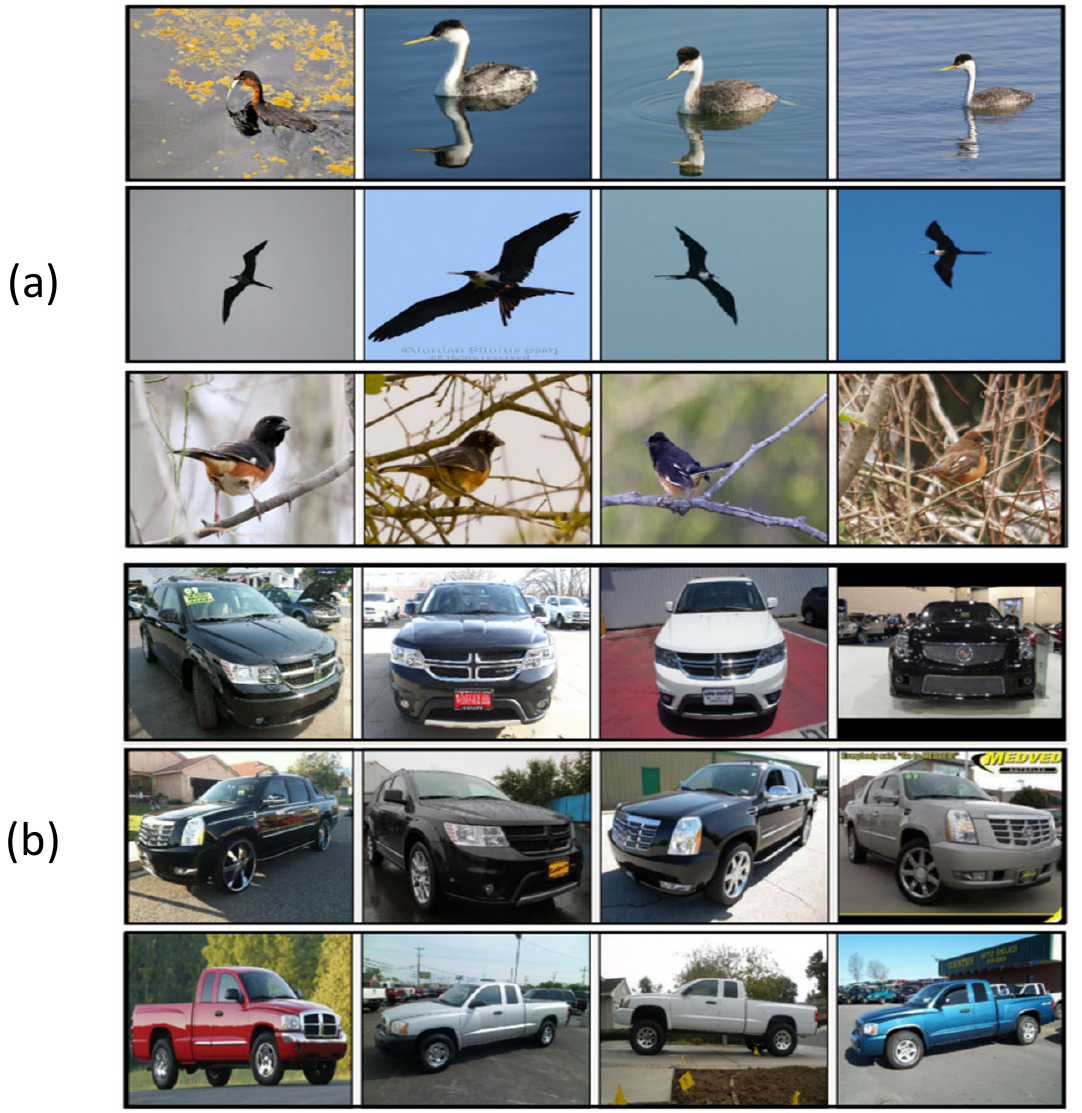}
\vspace{-1mm}
\caption{Top-4 neighbors of hierarchical proxies that are close to the boundary of the Poincar\`e ball on the CUB (a) and Cars (b) datasets.
The samples in each row are the nearest neighbors of a hierarchical proxy.
} 
\label{fig:sub_class}
\vspace{-5mm}
\end{figure}

In addition, to investigate the sub-class characteristics, we present the nearest neighbor samples of hierarchical proxies that are close to the boundary of the  Poincar\`e ball (\ie, representing a relatively low level of semantic hierarchy).
As shown in Figure~\ref{fig:sub_class}, each hierarchical proxy on the CUB dataset is associated with pose variation of birds and backgrounds of images (\eg, a bird floating on water, a bird flying in the sky, and a bird standing on a branch). 
On the Cars dataset, hierarchical proxies represent the viewpoints of the cars. (\eg, front, half side, and side views of the car).
These results suggest that HIER also captures sub-class relations, thereby enabling the model to learn intra-class variation and features shared across different classes. 

\subsection{Ablation Studies}
\noindent\textbf{Impact of HIER and hyperbolic space.}
We investigate the effect of HIER loss and hyperbolic space by comparing the performance of three models trained using two different metric learning losses as $\mathcal{L}_{\text{ML}}$ in Eq.~\eqref{eq:total_loss}: without HIER loss and with HIER loss in spherical space (\ie, HIER$_{sph}$), and with HIER loss in hyperbolic space (\ie, HIER (ours)). The results are summarized in Table~\ref{tab:ablation}. Overall, using the HIER loss leads to performance improvements across four datasets regardless of the kinds of metric learning loss. However, in spherical space, the improvement is not significant, possibly due to the limitations of spherical space in effectively representing the continuous latent semantic hierarchy of data discovered by the HIER loss. In contrast, applying the HIER loss in hyperbolic space results in significant performance improvements across all datasets. These results demonstrate that the HIER loss can effectively discover and deploy the latent semantic hierarchy of data in hyperbolic space.

\begin{figure} [!t]
\centering
\vspace{-1mm}
\includegraphics[width = 0.8
\columnwidth]{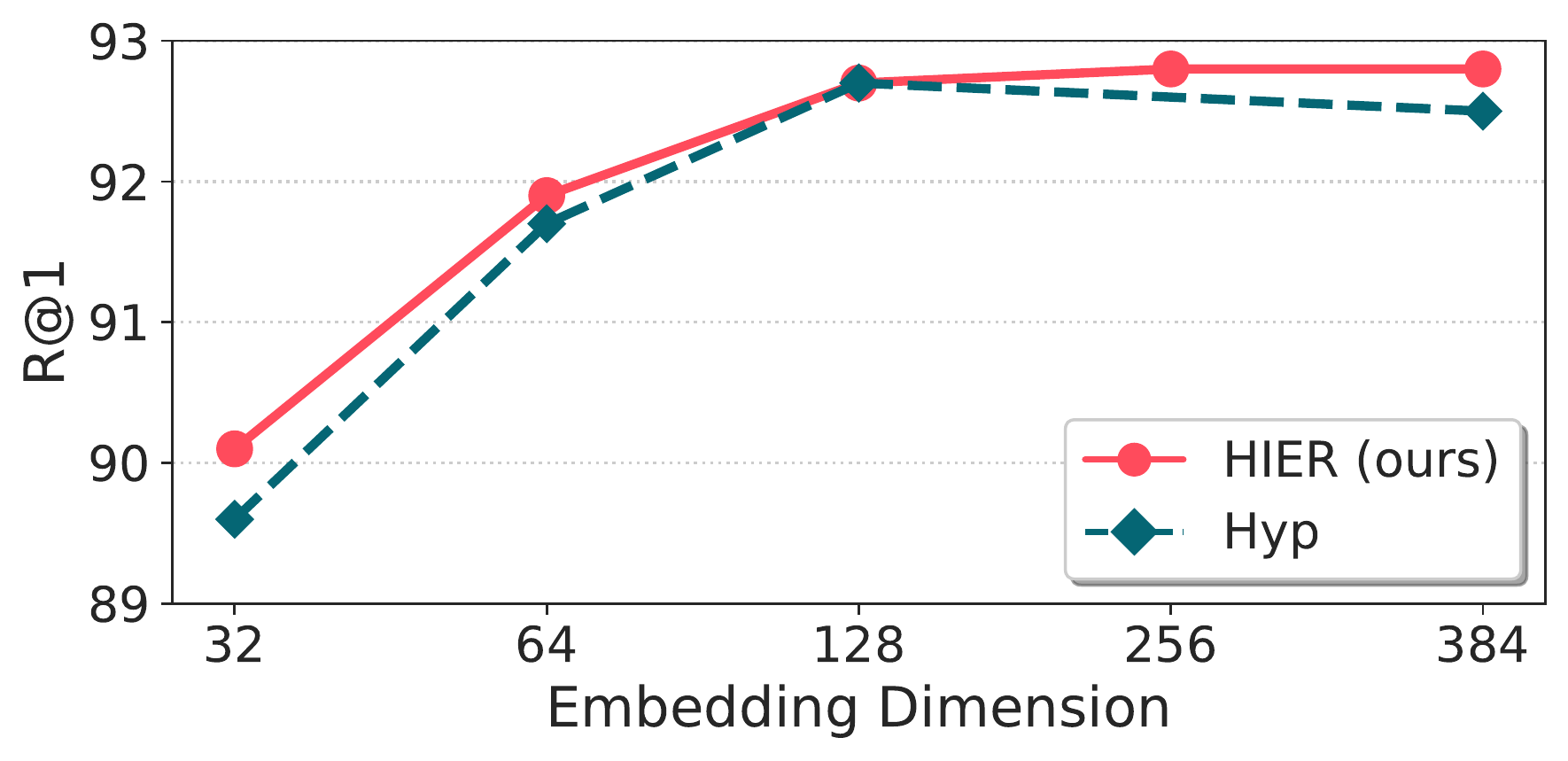}
\vspace{-2mm}
\caption{Comparison HIER and Hyp~\cite{ermolov2022hyperbolic} varying embedding dimension on the In-shop dataset using ViT as backbone network.
} 
\label{fig:embed_dim}
\vspace{-4mm}
\end{figure}

\vspace{1mm}
\noindent\textbf{Impact of embedding dimension.}
The embedding dimension is a seminal hyperparameter in image retrieval benchmarks due to the trade-off between efficacy and efficiency.
Therefore, we examine the effect of the dimension of embedding vectors in our method, compared to Hyp~\cite{ermolov2022hyperbolic}.
To this end, we evaluate our method on the In-shop~\cite{DeepFashion} dataset with ViT~\cite{ViT} with $d$-dimensional embedding vectors, where $d \in \{32, 64, 128, 256, 384 \}$.
Figure~\ref{fig:embed_dim} demonstrates that our method consistently outpaces Hyp with diverse embedding dimensions.
It is worth noting that the performance of our method increases consistently and stably according to the growth of the embedding dimension, while the performance of Hyp drops when the dimension is higher than 128.

\begin{figure} [!t]
\centering
\includegraphics[width = 0.96
\columnwidth]{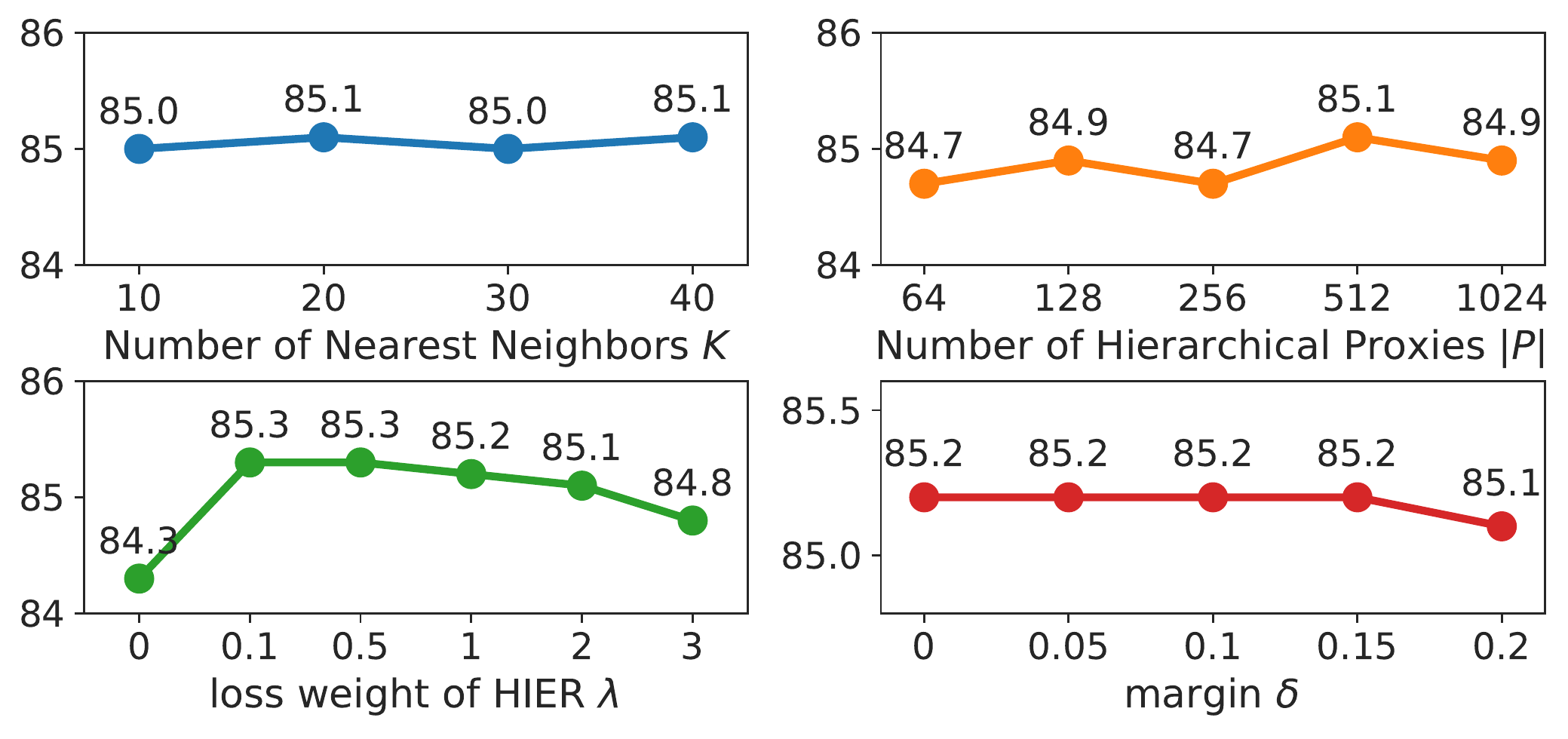}
\vspace{-1.5mm}
\caption{Recall@1 versus four hyperparameters of HIER on the Cars dataset using DeiT~\cite{DeiT} with 128 dimensions. Note that $\lambda=0$ denotes not using HIER loss.
} 
\label{fig:hyperparameter_ablation}
\vspace{-4mm}
\end{figure}

\noindent\textbf{Impact of hyperparameters.}
We investigate the impact of four hyperparameters on the performance of our method. These hyperparameters include the number of nearest neighbors $K$ in Eq.~\eqref{eq:triplet_condition}, the number of hierarchical proxies $|P|$, the loss weight of HIER $\lambda$ in Eq.~\eqref{eq:total_loss}, and the margin $\delta$ in Eq.~\eqref{eq:hc_loss}. We use the DeiT backbone and measure the performance on the Cars dataset. As shown in Figure~\ref{fig:hyperparameter_ablation}, the results demonstrate that remarkably robust to changes in these hyperparameters, suggesting that it can perform well regardless of their specific values. 


\begin{table}[!t]
\fontsize{8.2}{10.2}\selectfont
\centering
\begin{tabularx}{0.48 \textwidth}
{
  l
  >{\centering\arraybackslash}X
  >{\centering\arraybackslash}X
  >{\centering\arraybackslash}X
  >{\centering\arraybackslash}X
  }
 \toprule
\multicolumn{1}{c}{Methods}&
\multicolumn{1}{c}{CUB} & \multicolumn{1}{c}{Cars} & \multicolumn{1}{c}{SOP}& \multicolumn{1}{c}{In-Shop}\\ \midrule
PA &74.7 &84.3 &82.3 &90.4  \\
PA + HIER$_{sph}$ &75.1 &84.8 &82.4 & 90.6  \\
PA + HIER (ours)&75.2 &85.1 &82.5 &91.0 \\ \midrule
MS &75.4 &83.5 &80.0  &88.1  \\
MS + HIER$_{sph}$ &75.6&83.6 &80.0 &88.0 \\
MS + HIER (ours)&75.8 &84.3 &80.0  &88.2  \\
\bottomrule
\end{tabularx}
\vspace{-2.5mm}
\caption{Accuracy in Recall@1 of ours with two metric learning losses~\cite{kim2020proxy,wang2019multi}, and their variants on the four datasets. The network architecture is DeiT~\cite{DeiT} with 128 embedding dimensions. HIER$_{sph}$ denotes HIER over spherical space.}
\label{tab:ablation}
\vspace{-1mm}
\end{table}

\newpage
\section{Conclusion}
\label{sec:conclusion}
\vspace{-2mm}
In this paper, we have presented HIER, a regularization method for deep metric learning.
HIER discovers the latent semantic hierarchy of training data in a self-supervised learning fashion, and deploys the hierarchy to provide richer and more fine-grained supervision that the inter-class separability induced by common metric learning losses based on human-labeled classes. 
HIER consistently improve the performance of existing techniques when integrated with them, and consequently achieved the best records, surpassing even the prior art of hyperbolic metric learning.
Further, since it can be seamlessly incorporated with any existing metric learning losses, HIER enables taking full advantage of both recent hyperbolic embedding and the great legacy of hyperspherical embedding. 

\small{
\noindent \textbf{Acknowledgement.}
~This work was supported bythe NRF grant and  
the IITP grant 
funded by Ministry of Science and ICT, Korea
(NRF-2021R1A2C3012728--20\%,  
 IITP-2019-0-01906--10\%,     
 IITP-2020-0-00842--50\%,     
 IITP-2022-0-00926--20\%).    
 }

{\small
\bibliographystyle{ieee_fullname}
\bibliography{cvlab_kwak}
}

\newpage

\renewcommand\thesection{\Alph{section}}
\setcounter{section}{0}


\section{Appendix}
This supplementary material provides further analyses, additional experimental results, and implementation details 
that
are left out from the main paper due to the space limit.

\subsection{Effect of Deploying Semantic Hierarchy}
In this section, we investigate the effect of deploying semantic hierarchies by utilizing proxies. 
In particular, we quantitatively compare our method with  SoftTriple loss~\cite{Qian_2019_ICCV} which can model intra-class variance of data by utilizing multiple proxies per class. 
Table~\ref{tab:Proxy_comparison} shows the performance of two proxy-based methods and the number of proxies they used. 
In SoftTriple loss, $10$ proxies are assigned per class, and our method uses $512$ hierarchical proxies in addition to proxies for proxy anchor loss. 

The result shows that SoftTriple loss lags significantly in performance despite using a much larger amount of proxies compared to our method. 
The main difference between these methods is how they handle proxies. 
SoftTriple assigns multiple proxies per class, so their proxies can only represent sub-classes of data and model intra-class variance. 
On the other hand, proxies in our method can represent sub-classes or super-classes as well as predefined classes, thus allowing more flexibility in modeling more complex relations of data beyond intra-class variance.

\subsection{Embedding Space Visualization}
\begin{figure*}
    \centering
    \includegraphics[width = 0.99\textwidth]{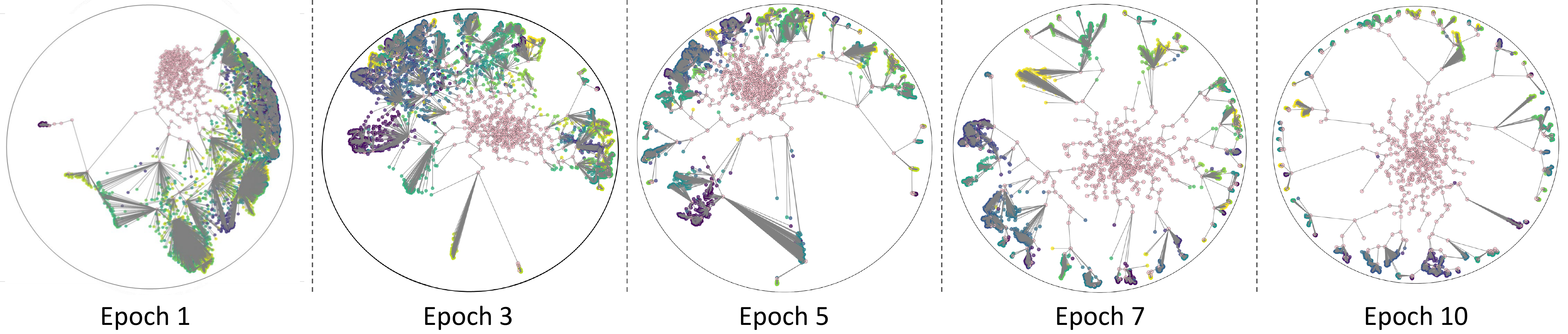}
    \caption{UMAP visualizations of our embedding space learned on the train split of the Cars dataset at different epochs. Pink points indicate hierarchical proxies, and other colors represent distinct classes. The gray line indicates the ancestor-descendant relation between the hierarchical proxy and data points.}
    \label{fig:umap_cars_epoch}
\end{figure*}

We further visualize the embedding space at the beginning of training, such as Epoch 1, 3, 5, 7, and 10.
The results of visualization presented in Figure~\ref{fig:umap_cars_epoch} show that the earlier embedding space does not construct a hierarchical structure between the embedding vectors and hierarchical proxies, while this structure is gradually constructed as the training epoch grows.
Specifically, as training progresses, 
the hierarchical proxies have ancestor-descendant relations between sample or other proxies, where these proxies can be regarded as predefined classes, sub-classes, and super-classes.
As a consequence, our method can discover the latent semantic hierarchy of training data, which provides rich and granular supervision beyond human-labeled classes. 

\subsection{More Qualitative Results}
\begin{figure*} [!t]
\centering
\includegraphics[width = 0.99\linewidth]{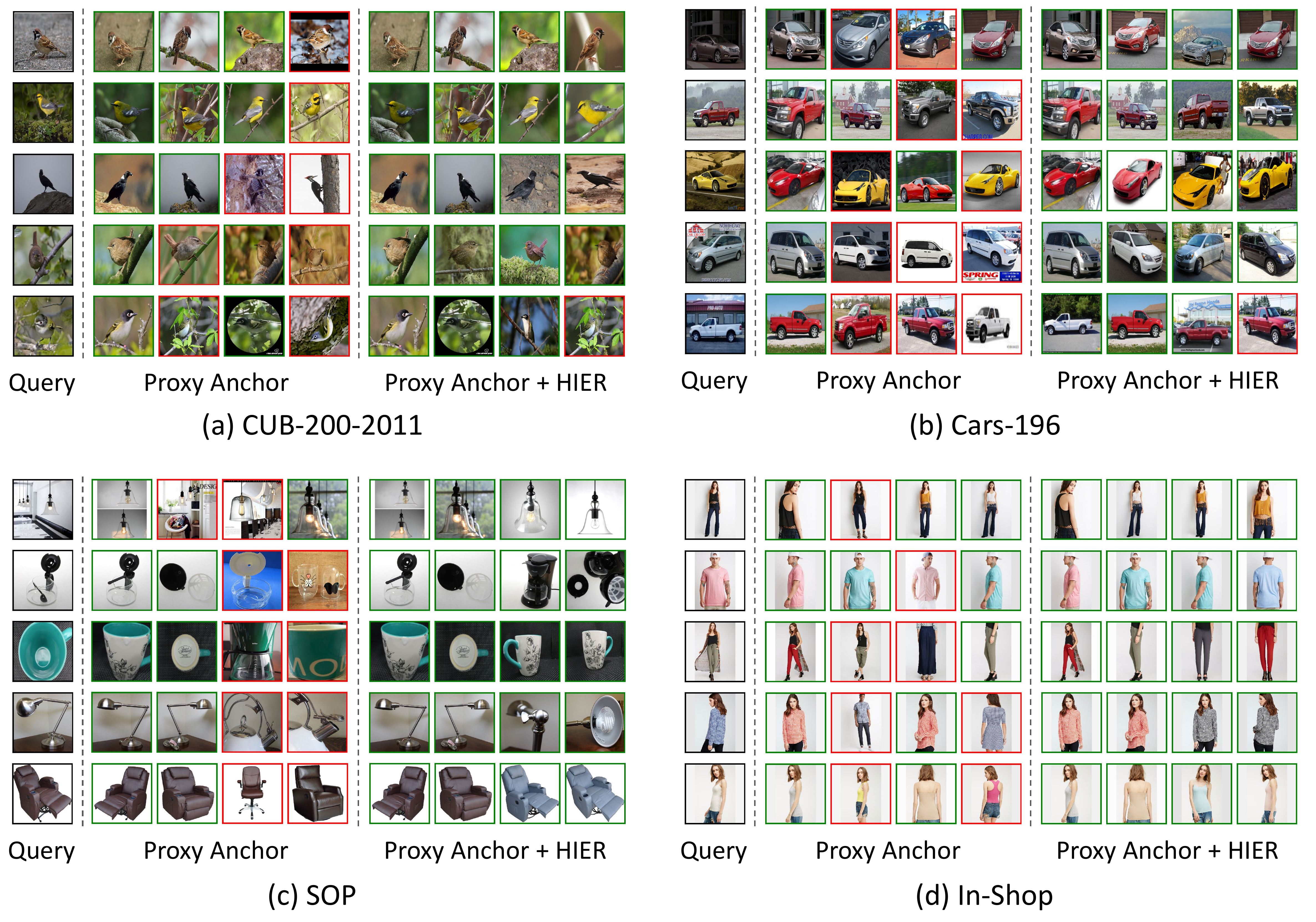}
\vspace{-1mm}
\caption{Qualitative results of ours and proxy anchor on the four public benchmark datasets, CUB (a), Cars (b), SOP (c), and In-Shop (d). Queries and the top 4 retrieval results of our method are presented. The true and false matches are colored in green and red, respectively.}
\label{fig:qual_all}
\end{figure*}

We further verify the effect of HIER in a qualitative perspective.
To this end, we present the qualitative results of our method, compared to those of a model optimized by only a metric learning loss, $\mathcal{L}_{\text{ML}}$ in Eq.~(9) of the main paper. 
We take proxy anchor loss~\cite{kim2020proxy} as the metric learning loss.
Figure~\ref{fig:qual_all} presents the qualitative results for the four public benchmark datasets, CUB~\cite{CUB200}, Cars~\cite{krause20133d}, SOP~\cite{songCVPR16} and In-Shop~\cite{DeepFashion}.
These results demonstrate that our method is robust against small inter-class variance (CUB and SOP), viewpoint variation and distinct color (Cars), and large intra-class variations and viewpoint changes (In-Shop) by discovering and deploying a latent semantic hierarchy of data, while the proxy anchor still suffers from those problems.

We note that all the results in the figure are obtained from the fully unseen class samples.
It implies that it allows the proposed hierarchical regularizer to discover the latent semantic hierarchy of data with no additional annotation for the hierarchy; our method allows the embedding space to be generalized well.

\begin{table}[t!]
    \centering
    \begin{tabular}{l|c c c c}
    \toprule
    Hyperparameters & CUB & Cars & SOP & In-Shop \\ \midrule
    total epochs &50 &50 &150 &150 \\
    warm-up epochs &1 &1 &5 &5 \\
    LR of the last layer &$\times1$ &$\times1$ &$\times10^2$ &$\times10^2$ \\ 
    weight decay &$\expnum{1}{2}$ &$\expnum{1}{2}$ &$\expnum{1}{4}$ &$\expnum{1}{4}$ \\ 
    margin $\delta$ &0.1 & 0.1 & 0.1 & 0.1 \\\bottomrule
    \end{tabular}
    \caption{Additional details of hyperparameters for network optimization. LR denotes the learning rate, and its value denotes how many times higher than the original learning rate.}
    \label{tab:implementation}
\end{table}

\begin{table*}[!t]
\fontsize{8.5}{10.5}\selectfont
\centering
\begin{tabularx}{0.99\textwidth}
{
  l
  >{\centering\arraybackslash}X
  >{\centering\arraybackslash}X
  >{\centering\arraybackslash}X 
  >{\centering\arraybackslash}X
  >{\centering\arraybackslash}X
  >{\centering\arraybackslash}X
  >{\centering\arraybackslash}X
  >{\centering\arraybackslash}X
  >{\centering\arraybackslash}X
  >{\centering\arraybackslash}X
  >{\centering\arraybackslash}X
  >{\centering\arraybackslash}X
  }
 \toprule
\multicolumn{1}{l}{\multirow{2}{*}[-3.5mm]{Methods}}&
\multicolumn{3}{c}{CUB} & \multicolumn{3}{c}{Cars} & \multicolumn{3}{c}{SOP}& \multicolumn{3}{c}{In-Shop}\\ 
\cmidrule(lr){2-4} \cmidrule(lr){5-7} \cmidrule(lr){8-10} \cmidrule(lr){11-13} 
&\#Proxies &R@1 &R@2 &\#Proxies &R@1 &R@2 &\#Proxies &R@1 &R@10 &\#Proxies &R@1 &R@10 \\ \midrule
SoftTriple~\cite{Qian_2019_ICCV} &\textbf{1,000} &72.7 &82.7 &\textbf{980} &83.2 &90.2 &\textbf{113,180} &80.9 &91.3 &\textbf{39,970} &88.5 &97.3 \\
PA + HIER (ours)&612 & \textbf{75.2} &\textbf{84.2}& 610 &\textbf{85.1} &\textbf{91.2} &11,830 &\textbf{82.5} &\textbf{92.7} &4,509  &\textbf{91.0} &\textbf{98.0} \\ 
\bottomrule
\end{tabularx}
\vspace{-1mm}
\caption{Comparison between methods using proxies in terms of performance and the number of proxies on the four benchmark datasets. In these experiments, the backbone network is initialized by weights of DeiT~\cite{DeiT}.}
\label{tab:Proxy_comparison}
\vspace{-1mm}
\end{table*}

\subsection{Additional Implementation Details}
Since each dataset that we utilize to evaluate our method has a different characteristic, we set different hyperparameters for network optimization according to each dataset.
The summary of the settings of hyperparameter for four datasets are presented in Table~\ref{tab:implementation}.
For all backbone network variants (\ie, ViT-S~\cite{ViT}, DeiT-S~\cite{DeiT}, and DINO~\cite{DINO}), our model is trained with 50 epochs on CUB~\cite{CUB200} and Cars~\cite{krause20133d}, and 150 epochs on SOP~\cite{songCVPR16} and In-Shop~\cite{DeepFashion}.
As a warm-up strategy, the last layer which consists of a linear layer followed by the exponential mapping layer is only trained, while the pretrained backbone network is not updated; the warm-up strategy is applied for 1 epoch on CUB and Cars, and 5 epochs on SOP and In-Shop.
On the other hand, we use a high learning rate for the last layer~(\ie, embedding layer) by scaling 10$^2$ times for SOP and In-Shop.
Furthermore, we set the weight decay factor as $\expnum{1}{2}$ for CUB and Cars, and $\expnum{1}{4}$ for SOP and In-Shop. In all experiments, the margin parameter $\delta$ of HIER loss in Eq.~(8) is fixed at $0.1$.

\end{document}